\documentclass{article}

\usepackage{microtype}
\usepackage{graphicx}
\usepackage{subfigure}
\usepackage{booktabs} 
\usepackage{epstopdf}

\usepackage{hyperref}
\usepackage{lipsum}
\usepackage{graphicx}

\title{Learning from Higher-Layer Feature Visualizations}

\usepackage{amsmath,amsfonts,amssymb,amsthm,epsfig,epstopdf,titling,url,array}
\theoremstyle{plain}
\newtheorem{defn}{Definition}[section]

\author{
  Nikolaidis, Konstantinos\\
  University of Oslo\\
  \texttt{konstan@ifi.uio.no}\\\\
  \and
  Kristiansen, Stein\\
  University of Oslo\\
  \texttt{steikr@ifi.uio.no}\\\\
 \and
 Goebel, Vera\\
  University of Oslo\\
  \texttt{goebel@ifi.uio.no} \\\\
  \and
  Plagemann, Thomas\\
   University of Oslo\\
  \texttt{plageman@ifi.uio.no}\\\\
}

\begin{document}

\maketitle

\begin{abstract}
Driven by the goal to enable sleep apnea monitoring and machine learning-based detection at home with small mobile devices,  we investigate  whether interpretation-based indirect knowledge transfer  can be used to create classifiers  with acceptable performance.  Interpretation-based indirect knowledge transfer means that a classifier (student) learns from a synthetic dataset based on the knowledge representation from an already trained Deep Network (teacher). We use activation maximization to generate visualizations and  create a synthetic dataset to train the student classifier. This approach has the advantage that student classifiers can be trained without access to the original training data. With experiments we investigate the feasibility of interpretation-based indirect knowledge transfer and its limitations.  The student achieves an accuracy of 97.8\% on MNIST  (teacher accuracy: 99.3\%) with a similar smaller architecture to that of the teacher. The student classifier achieves an accuracy of 86.1\% and 89.5\% for a subset of the Apnea-ECG dataset (teacher: 89.5\% and 91.1\%, respectively).

\end{abstract}
\section{Introduction}
\label{intro}
In  our project we aim to enable Obstructive Sleep Apnea (OSA) monitoring and detection at home with consumer electronics. Machine learning has shown to be rather efficient for OSA detection, e.g., \cite{uddin2018classification}. To be able to run classifiers on mobile computing devices with restricted resources, e.g., smart phone or smart watch, we need to develop an approach to create  small classifiers with high performance. 

Literature suggests that larger deeper neural networks  outperform smaller shallower ones on large, challenging datasets \cite{he2016deep,howard2017mobilenets,goodfellow2016deep}. This insight triggered the idea to leverage the knowledge representation of deep neural networks (DNN) to train the small classifier, i.e., to use knowledge transfer to improve its performance. There exist many knowledge transfer techniques (see Section \ref{related_work}). However, the challenge we address is different from existing approaches, because we want to transfer knowledge from a larger classifier (teacher) to a smaller classifier (student) and we want to do this independently of learning method and without access to the original training data. The latter is due to the many issues with sharing health data of patients (in our case sleep monitoring data).

The essence of the knowledge transfer approach presented in this paper is to re-use recent visualization-based results for interpretable machine learning with a new goal. Instead of using visualizations to enable a human expert to understand the internals of a model, we use the visualization approach to create a synthetic dataset to train the student classifier. Therefore, we call this approach interpretation-based indirect knowledge transfer. We take advantage of the fact that neuronal activation is generally multi-faceted \cite{nguyen2016multifaceted} to create a synthetic dataset of visualization with the diversity that has the capability to train a student classifier. Our experimental work reveals that our goal that a student classifier trained on the synthetic dataset achieves  better performance than if it would have trained on the original training set cannot be met. However, we show  that the larger the synthetic dataset is, the better the performance of the student classifier such that it is possible to come close to this goal using the proposed technique.

We analyze the impact of the architectural similarity of student and teacher and  investigate the performance of interpretation-based knowledge transfer between different learning methods, i.e., from a Convolutional Neural Network (CNN) to Support Vector Machines (SVM), Random Forest (RF), and a Multilayer Perception architecture (MLP). Our results show that the more similar student and teacher are  architecturally and algorithmically, the higher the performance of the student trained on the visualizations.

We introduce two new metrics to measure the difficulty that other trained classifiers  have in identifying the synthesized data, and whether we can pass "hidden" messages from a classifier to another without the other classifiers (the majority of them) realizing it. Based on these metrics, we evaluate the performance of the synthesized dataset in comparison to the real dataset. The other trained classifiers are either CNN or humans.
 
\textbf{ Our Contributions}: (1) We demonstrate that a classifier can learn from the feature visualizations of another classifier. (2) We develop a novel technique of knowledge transfer that does not require the training data for the student to train and is independent of architecture or algorithm for the student network. It makes the student generalize with various degrees of success to the real data depending on the architectural or algorithmical proximity between student and teacher and the generating algorithm. (3) Two new metrics that give quantitative measures for difficulty in interpreting the visualizations applicable to any point in the feature space.

The remainder of this paper is structured as follows. In Section \ref{related_work} we present a more detailed analysis of related work. In Section \ref{method} we explain the design of our method and we evaluate the proposed technique in Section \ref{eval}.  In Section \ref{concl} we  present conclusions and  future work.

\section{Related Work}
\label{related_work}

 Recently, many new techniques for transfering knowledge have been proposed, especially with the goal to reduce the size of a  DNN  to decrease the execution time and reduce memory consumption. Existing model compression techniques, e.g., via pruning or parameter sharing \cite{cheng2017survey,hassibi1993second,han2015learning,srinivas2015data} can be considered as a form of knowledge transfer from a trained teacher to a student. Other types of methods, transfer the knowledge from a smaller to a larger DNN to make it learn faster \cite{chen2015net2net} or even between different task domains \cite{pan2010survey,yim2017gift}.

 In  the knowledge distillation method \cite{luo2016face}, the student network is trained to match  the softmax output layer logits of the trained teacher network and  the classes of the original data. \cite{romero2014fitnets} introduce fitnets, an extension of the knowledge distillation method to train thinner deeper networks (student) from wider shallower ones (teacher).   \cite{buciluǎ2006model} investigate the compression of large ensembles (like RF, bagged decision trees, etc.) via the use of a very small artificial neural network (ANN). As a universal approximator the ANN  is able to generalize to mimic the learned  function of the ensemble  given sufficient data. To train the ANN they create a larger synthetic dataset based on the real dataset that is labeled by the ensemble.   \cite{luo2016face} use knowledge distillation  on a selection of informative neurons of top hidden layers to train the student network. The selection is done by minimizing an energy function that penalizes high correlation and low discriminativeness.

It is worth noting that our technique differs from the above since we do not aim to perform any form of knowledge distillation. Our aim is for the student to independently train on a synthetic dataset created by capturing learned high-layer features of the teacher. For this reason, the original data is not needed to train the student. Our technique is additionally applicable to different learning algorithms.

Regarding the understanding of internal representations of a DNN, \cite{erhan2009visualizing} introduce the activation maximization (AM) technique, among others, for qualitative evaluation of higher-level representations of the internal representations of two unsupervised  deep architectures. \cite{nguyen2015deep} generate fooling images either directly or indirectly (via a compositional pattern producing network) encoded via maximizing the output layer of a DNN. They show that the images produced can be unrecognizable for a human observer.  In their follow up work,  \cite{nguyen2016synthesizing} use a deep generative network  to synthesize images that maximize the output of a neuron of a certain layer of the network. We base our approach on some of these insights.

\section{ Method}
\label{method}

In this section, we explain intepretation-based indirect knowledge transfer from a more general perspective, present our proposed architecture, and specify the procedure we use to generate the synthetic dataset.

\begin{figure*}[ht]
\centering
\includegraphics[scale=0.33]{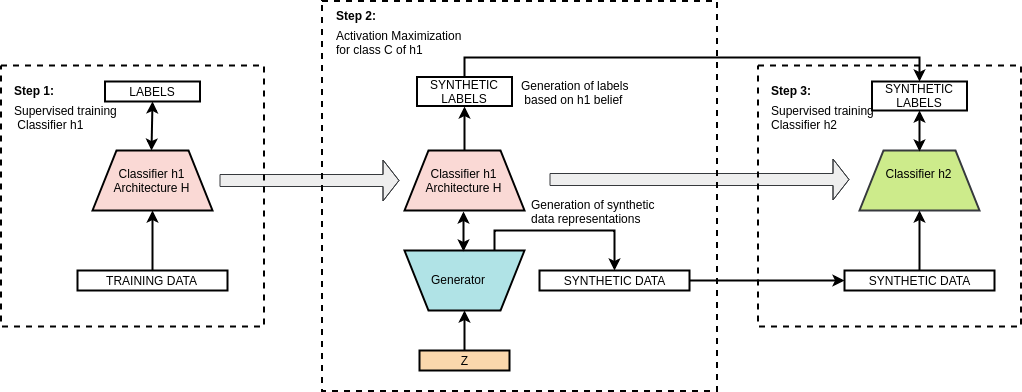}
\caption{\textit{ The Generating procedure we follow includes AM via using a generator  network G similarly to \cite{nguyen2016synthesizing}. }}
\label{fig:Method1}
\end{figure*}
\subsection{ Intepretation-Based Indirect Knowledge Transfer }
\label{method_1}

We are interested in transferring the knowledge of a given trained DNN  $h_1$ to another model or learning algorithm $h_2$. We assume that the original data with which $h_1$ is trained is not available for  training $h_2$. We aim to enable $h_2$ to classify data that come from the same distribution as $D$ with a similar performance as $h_1$. The only way to achieve this without access to $D$ is to extract the knowledge accumulated by $h_1$. We perform this by generating visualizations that correspond to strong beliefs of what $h_1$ assumes a class is. These visualizations will then be used as a synthetic  dataset to train $h_2$.

 We define \textit{intepretation-based indirect knowledge transfer} for a classifier $h_2$ to mean the following: a model $h_1$ is first trained on  $D$ with distribution $p_{Data}$. Then a generation procedure $Gen\{h_1\}$ produces a dataset $D_2$ to train classifier $h_2$.

Depending on the success of $Gen\{h_1\}$ to map the  important features from the task domain learned  by $h_1$ into $D_2$, and the algorithmic and architectural similarity between $h_1$ and $h_2$, we show that it is possible for $h_2$ to learn to perform the classification task $h_1$ has learned.

\subsection{Design} 
\label{method_2}

Our proposed design (see Figure \ref{fig:Method1}) is based on three basic steps:

 Step 1 - Training of the teacher. We train the teacher network $h_1$ in a supervised  manner with  $D$  to learn the underlying data distribution $p_{Data}$. This requires the original labeled training data.

 Step 2 - Creating the synthetic dataset. To enable indirect knowledge transfer, we create a synthetic dataset $D_2$ that captures features that $h_1$ has learned from  training on $D$.   $Gen\{h_1\}$ is used to create this synthetic dataset, and is one of the core elements of the method. In $Gen\{h_1\}$,  we perform AM  via a deep generator network (G)  that transforms a small noise vector Z to  examples that strongly activate a predefined neuron   (see Section \ref{method_3} for further details).  Inspirations for this design were \cite{baluja2017adversarial,nguyen2016synthesizing}. After the synthetic set is created we create its labels. If $\mathbf{x_s}$ corresponds to a synthetic example created by  $Gen\{h_1\}$, we give $\mathbf{x_s}$ the label that $h_1$ chooses for it, i.e., either the class with the maximum output probability, $\arg\max_i \{h_{1_{iL}}(\theta,\mathbf{x_s}) \}$
where $i$ corresponds to the possible class of the output, L denotes the output layer of $h_1$ and $\theta$ corresponds to the parameter vector of $h_1$, or  the softmax of the output to better capture the output probabilities of $h_1$.

 Step 3 - Training of the student.  As final step the student $h_2$ is trained via the synthetic data and labels produced by step 2. $h_2$ can be smaller or larger DNN than the  $h_1$, or even be based on a different learning method.

The choice of $Gen\{h_1\}$ and which method to use for $h_2$ are two central decisions. Next, we discuss  the choice of $Gen\{h_1\}$ and in Section \ref{eval} we evaluate the performance of different $h_2$ methods.

\subsection{ Generation }
\label{method_3}

For  knowledge extraction and generation we consider two   visualization approaches: AM and code inversion \cite{mahendran2016visualizing}. However, code inversion requires the original training data, or the logits of the data from the fully connected layer we will try to match. This defeats our goal to train $h_2$ without access to $D$. Therefore, we use AM for the generation of the synthetic dataset. Contrary to other works like  \cite{nguyen2016synthesizing} the goal of $Gen\{h_1\}$ is not to produce realistic looking synthetic data, but instead to investigate whether we can reliably train $h_2$ from the synthetic dataset and how sensitive the procedure is to the choice of  the architecture or learning method of $h_2$.

To synthesize data that the trained model $h_1(\theta,x)$ perceives as of class $i$,  $Gen\{h_1\}$  uses AM \cite{erhan2009visualizing} on the activations of output layer L of  $h_1(\theta,x)$ via G such that:
\begin{equation}
\theta^*_G=\arg\max_{\theta_G  s.t ||G(\theta_G,\mathbf{Z})||\leq R } \{h_{1_{iL}}(\theta,G(\theta_G,\mathbf{Z}))\}
\end{equation}

 where $i$  corresponds to the class we want to find a visualization for. $\theta$ is the parameter vector of $h_1$, and $\theta_G$ the parameter vector of  G. Note that $\theta$ is static because $h_1$ is already trained and we are optimizing for $\theta_G$ given varying "pseudo inputs" of $\mathbf{Z}$. $R$ is the maximum value of the norm of the output of G for a given norm (we use sigmoids on the output of G), and $L$ denotes the output layer. We choose to perform AM in the output layer of $h_1$ since we want  visualizations that correspond to a strong belief for a class in $h_1$.  We use $\mathbf{Z}$ as an input random noise vector (mainly from a  uniform distribution) which G transforms into the visualization. We intentionally do not give any additional priors to  $Gen\{h_1\}$, since we do not want to constrain our exploration of the movement in the feature space.
 
 We stop the AM when the output of the target class neuron $i$ is higher than all the other ouput neurons and exceeds a threshold T:

\begin{equation}
h_{1_{iL}}(\theta,G(Z, \theta_G^*))>T
\end{equation}

This implies that the space that satisfies Eq.(2) is a subspace of the  space of class $i$, as defined by its decision boundaries. We  assume that softmax activation is used on the output layer.

\par We specify the the loss of G to maximize the logits of each class of the classifier ($h_1$ in Figure \ref{fig:Method1}). We then repeat the procedure for a number of iterations, and end up with a synthetic dataset comprised of the visualizations of $h_1$ for all of the classes.

\subsubsection{ Reinitializations }

\par The goal of  $Gen\{h_1\}$ is to generate  visualizations such that $h_2$ can learn and generalize to D. We therefore need to generate many visualizations of each class. However, with the procedure above we get data from the same limited  neighborhood of the feature space, depending   on the variation of the input  noise Z in  G.
To ensure that we  capture the space of  a class defined by $h_1$ better, we reinitialize  $\theta_G$ and repeat the procedure  a certain number of times until there is no additional improvement on the generalization capability of $h_2$. We test this by measuring $h_2$ accuracy on the test data. We find that repeated initializations of $\theta_G$ with the default truncated normal initializer almost always yield the same class for $h_1$. A test with  1000 reinitializations of $\theta_G$ resulted in, $argmax_c\{h_1(\theta,G(\theta_G,z))\}$  always having the same class. This indicates that  the initial subspace covered by  the output of G is constrained to a subspace of the space of this class.

 Due to the small default weight initializations in relation to the total space, the potential starting points for the output of G are generally confined. Additionally, AM is hindered when we increase the initial weights to relatively large values (by varying the truncated initializer). This might be attributed to very large values for the generated images (output of G) that are very close to 0 or 1 which have a negative effect to the learning procedure. This problem is also related to the variance or limits of $\mathbf{Z}$. However, we  cannot increase this variance without limit, as this also hinders learning (a higher variance  implies inconsistently large numbers, which leads to stronger activations). In summary, potentially we  cannot  capture all of the important parts of $p_{Data}$ that $h_1$ has learned due to  limitations in the randomness of the  initial $\theta_G$ and $\mathbf{Z}$, depending on the loss landscape. We therefore perform an initial step of pseudorandom  movement in the feature space (i.e., output space of G and input of $h_1$) before the actual AM. 

\subsubsection{ Attraction-Repulsion for diversity}

\begin{figure}
\centering
\includegraphics[scale=0.25]{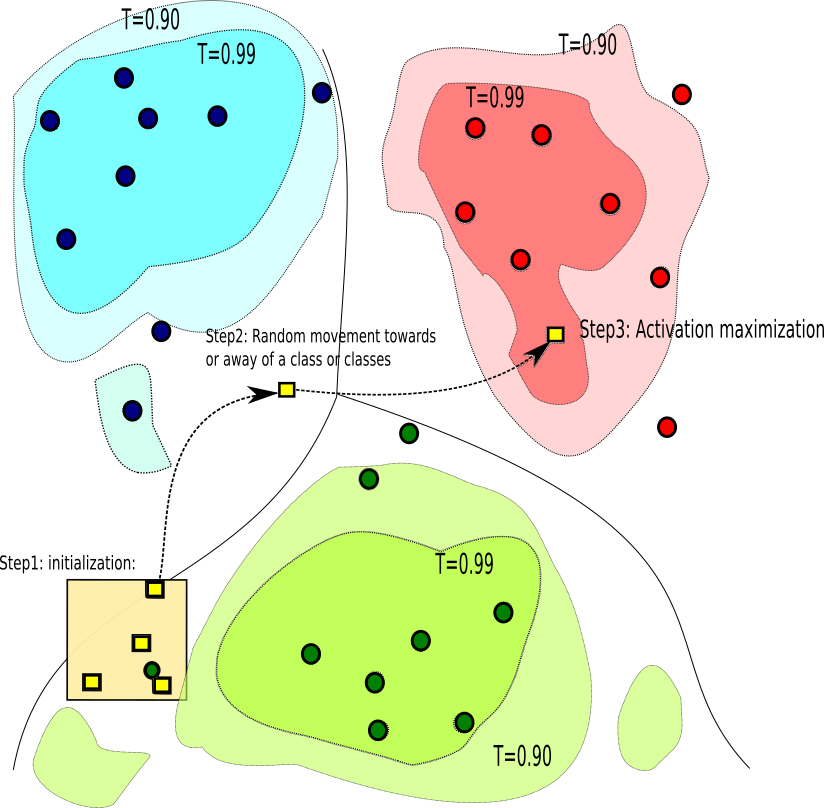}
\caption{\textit{ Example of the generation procedure. }}
\label{fig:explan1}
\vskip -0.2in
\end{figure}

\par  We perform pseudorandom movements via optimizing $\theta_G$ to either maximize or minimize the $h_1$ activation  for a random subset of classes. This is performed across  a random number of steps.
This pre-AM step changes  the initial position of the output of G in the feature space as defined by $\theta_G$. If  the change in the initial position is sufficient,  the entering  to the subspace which satisfies our AM threshold from Eq.(2) can potentially  capture a different facet of the strong output neuronal activation. In Figure  \ref{fig:explan1} we show an example. The green, red, and blue dots depict data points. The  green, red, and blue areas depict areas of the feature space that satisfy Eq.(2) for the maximum  logit class neuron of $h_1$. The lighter regions depict areas where threshold T is smaller and thus are supersets of the previous. The yellow rectangles depict points of  G with different parameter tunings during Steps 1-3. In Step 1 after the initialization the position is defined from  the randomized vector $\theta_G$ and from  $\mathbf{Z}$. Since both are confined, the output of G will also be confined ($l_\infty$ hypershere). In Step 2 we execute the pseudorandom movement. In Step 3  AM towards the  specified class (red) is performed.

\section{ Evaluation}
\label{eval}

In this section, we investigate the viability of the proposed approach. We perform three experiments. Experiment 1 (Section \ref{eval_1}) acts as an empirical proof of concept. We examine the performance of different student $h_2$ architectures and methods on the MNIST dataset \cite{lecun1998gradient}. As $h_1$ we use a small CNN and as G a small deconvolutional deep generator. Since we are interested in studying the impact of different architectures and learning methods we do not use existing pre-trained classifiers, e.g., LeNet \cite{lecun1998gradient}.

 We study  in Experiment 1 a modification of the proposed method: we specify the additional goal on G, to maximize the $l_2$ distance from the real data (for this experiment we access the original data) in the second to last hidden layer of $h_1$, to instigate visualizations with high level features that are not in D. We investigate whether we can learn with this added constraint.
 
 The last part of Experiment 1 serves as a basis for Experiment 2 (Section \ref{eval_2}). In Experiment 2 we examine the ability of a set S of potential learning methods  to predict $h_1$ beliefs for a  given set of visualizations. Based on this, we define properties to measure the absolute and relative difficulty of the set S in performing this task. We also evaluate this on the basis of each classifier of S independently.

 In Experiments 1 and 2, we demonstrate that: (1) $h_2$ can learn from the visualizations albeit to varying degrees of success depending on  the learning method and (2) that the visualizations are interpetable using a third method (or set of methods) to varying degrees of success  depending on the architectural and algorithmic similarity with $h_1$. In Experiment 3, we use a realistic healthcare application to evaluate both aspects of our approach  simultaneously (Section \ref{eval_3}).

\subsection{Proof of Concept with MNIST Dataset}
\label{eval_1}

For  $h_1$ we use a small CNN with dropout (conv-maxpool-conv-maxpool-fc1-dropout-fc2-sigmoid). We use seven 
 different $h_2$: (1) A CNN that has exactly the same architecture as $h_1$ (CNN$_{ID}$). (2) A network that has similar architecture as $h_1$, but is  smaller with half  of the channels for all the convolutional layers and half of the neurons in the fully connected layers  (CNN$_S$). (3) A network that is larger than $h_1$, with one convolutional and one fully connected layer more, and  that has different pooling and activation functions, i.e., average instead of max for pooling and elu instead of relu for activations (CNN$_L$ ). (4) A convolutional network with even less weights than CNN$_S$ (CNN$_s$). (5)  A MLP architecture with a small perceptron with three hidden layers and relu activations. (6) A simple SVM.  (7) A RF with 100 trees.

\par We test these classifiers  as $h_2$ on the test set of the MNIST dataset. Table \ref{table:Res_ev1} presents the classification results for the baseline (when $h_2$ is trained with the real data) and the synthetic dataset. Additionally, we experiment with soft labels and  randomized constrained T. Our best results are: an accuracy of 98.13$\pm$0.03\% for CNN$_{ID}$ and 97.84$\pm$0.06\% for CNN$_S$  (see Appendix).

\begin{table}
\caption{Results for 60000 visualizations. T=0.99, hard labels}
\label{table:Res_ev1}
\vskip 0.15in
\begin{center}
\begin{small}
\begin{sc}

\begin{tabular}{lcr}
\toprule
  $h_2$ Acc(\%) &Base&$h_1$Vis.\\ 
\midrule
 CNN$_{ID}$ &99.30$\pm$0.03 &95.51$\pm$0.28\\
 MLP&94.80$\pm$0.10&48.91$\pm$0.18\\
 RF&92.12$\pm$0.10&23.20$\pm$1.40\\
 SVM&94.04$\pm$0.00&39.71$\pm$0.00\\
 CNN$_S$&99.23$\pm$0.03&95.11$\pm$0.33\\
 CNN$_L$&99.42$\pm$0.03&94.53$\pm$0.41\\
 CNN$_s$&98.89$\pm$0.03&91.80$\pm$0.30\\
\bottomrule
\end{tabular}%
\end{sc}
\end{small}
\end{center}
\vskip -0.1in
\end{table}

The results in Table \ref{table:Res_ev1} show a clear trend that when $h_2$ has the same or similar CNN architecture like $h_1$, it is able to learn from the synthetic visualizations and generalize relatively efficiently to the real test data. However, we  observe a significantly lower performance for the other classifiers, i.e.,  MLP, RF, and SVM. Since all clasifiers are able to succesfully learn the visualization classes during training the lower performance is caused by a lower generalization capability. However, most of the algorithms are generalizing better than random choice.

\par For simpler algorithms like SVMs we can give a potential explanation for the low performance. As shown in Figure \ref{fig:explan1} and  in \cite{nguyen2015deep}, inside the class boundaries of a trained model (in our case $h_1$),  exist regions of high confidence that are sparse in the sense that they do not contain many data points from the training data.  Since we try to capture these high confidence regions from different directions (via the randomized repeated initializations), there can be cases where (1) areas that are not in the original  data distribution are captured and (2) areas that are in the  data distribution, are not captured sufficiently by the synthetic dataset. For the SVM, the final decision hyperplane after the training on the vizualizations can be very different from the original, because it works with points that are near the boundaries (the support vectors), and these boundary data points  can be very different in the synthetic dataset due to the aforementioned reasons. 

Images from this procedure for MNIST can be found in the Appendix.
In the following sections we investigate how $h_2$ performance changes as we increase the number of visualizations synthesized.

\subsubsection{Size of the synthetic dataset}
\label{eval1_1}

We evaluate the performance of $h_2$ with increasing  size of the synthetic dataset. As reference, we measure the  performance  of  $h_2$ with subsets of the original training data with the same size as the synthetic dataset. $h_2$ has exactly the same architecture as $h_1$ in this experiment.

\begin{table}
\vskip -0.2in

\caption{Accuracy vs size of the synthetic dataset.}
\label{data_incr}
\vskip 0.15in
\begin{center}
\begin{small}
\begin{sc}
\begin{tabular}{lcccr}
\toprule
Data size(\#) & Acc Synth & Acc Real \\
\midrule
400     &63.0$\pm$2.0& 92.5$\pm$0.2  \\
2000    & 85.5$\pm$1.1&96.8$\pm$0.1 \\
4000    & 87.5$\pm$1.1&98.3$\pm$0.02 \\
12000   & 92.7$\pm$ 0.5&98.8$\pm$0.02 \\
20000   & 94.0$\pm$0.1&99.1$\pm$0.05 \\
40000   & 94.8$\pm$0.2&99.3 $\pm$0.01\\
60000   &95.5$\pm$0.3 & 99.3$\pm$0.02       \\
80000   &95.7$\pm$0.2 & -\\
\bottomrule
\end{tabular}
\end{sc}
\end{small}
\end{center}
\vskip -0.2in
\end{table}
 Table \ref{data_incr} shows that increasing the size of the synthetic dataset  has a positive effect on the performance of $h_2$.  As expected, training with visualizations is much less efficient than training with the real data. For any given accuracy, we need more synthetic  data than original  data. On the other hand, the synthetic dataset can be arbitrarily large, which means that the performance of $h_2$, as long as $h_2$ has the capacity, could match that of $h_1$. This also depends on $Gen\{h_1\}$, learning the important features learned from $h_1$.

\subsubsection{Avoiding High Level Features of D}
\label{eval1_2}

We evaluate how well $h_2$ can train with visualizations that are intentionally dissimilar from the real data in terms of the features captured by high-layer neurons that are not output neurons. Inspired by \cite{reddy2018nag} and \cite{liu2016coupled}, and knowing that  higher-layer neurons map predominantly high-level semantics  \cite{erhan2009visualizing}, for a given distance metric, we maximize the distance between the logits for real and generated data for an intermediate  fully connected layer:

\begin{equation}
Loss_2=-\sum_{n=0}^{B_s} d(h_{1_{L-1}}(\theta,\mathbf{x}_n),h_{1_{L-1}}(\theta,G(\theta_G,\mathbf{z})))
\end{equation}

where all $\mathbf{x}_n\in D_i$ where $D_i$  is the subset of $D$ with labels of class $i$ assuming that we perform AM for class $i$. For simplicity we choose $l_2$ as distance metric $d$. $B_s$ is the size of the batch and $L-1$ denotes the second to last layer. Therefore, we want to capture features that activate maximally the output (class) neuron, but are far from the ones of the real data in the previous layer. During training of G we perform this update (i.e., movement) and AM in separate steps, and we use a learning rate that is $1/3-1/5$ of that of the learning rate of AM depending on the  pseudorandom movement (see Section \ref{method_3} ).

After the synthesis of the visualizations with the extra condition, we  train a  $h_2$ with the same architecture  as $h_1$ but with different weight initializations.  We get an accuracy of 94.33$\pm$0.12 for a synthetic dataset with 60000 images, while the original procedure achieves  95.51\% accuracy.

\subsection{Difficulty of Interpretation} 
\label{eval_2}

 Next, we investigate how difficult it is for classifiers with different architectures to find the class which an example $\mathbf{x}_s$ (visualization for our experiments) is classified by $h_1$. To measure the total difficulty of a set S of different methods, we define two properties: the fooling example and hidden message. We use these properties to evaluate how difficult it is for different sets of classifiers to identify visualizations produced using the CIFAR10 dataset.

\subsubsection{Property Definitions}
\label{Eval_2_2}

We define two properties of points in the feature space to quantify how difficult if is for a set of  classifiers $S$ to identify  the decision of $h_1$ for a given example $\mathbf{x_s}$. The fooling example property provides a direct estimate, whereas the hidden message property provides a relative estimate in relation to another student network $h_2$.

 In the following definitions, let $S=\{h_1,h_2...h_N \}$ with $h_i \in H$, $h_i:X \rightarrow Y,i=1...N$, where H is the space of  possible classifiers   and $|S|=N$. For a given metric $m(h_j,D) \in [0,1]$
, a dataset  $D_T=\{(\mathbf{x}_1,\mathbf{y}_1),(\mathbf{x}_2,\mathbf{y}_2),...,(\mathbf{x}_M,\mathbf{y}_M) \}$ with $\mathbf{x}_1,...\mathbf{x}_N, \in X$  and a  given threshold $A \in [0,1]$
,we have that  $m(h_i,D_T)>A$  $\forall$ $h_i \in S$ (assuming higher m means better performance). First, we redefine the fooling examples from \cite{nguyen2015deep}  using the notion of a comparative set S:
 
\begin{defn}
 Fooling example:
Given classifier $h\notin S$, we define a fooling example   for S  to be  a $\mathbf{x}_s \in X$ such that, $h(\mathbf{x}_s)=l\in Y$, and there exist a subset of S such that $S_c=\{h_i \in S | h_i(\mathbf{x})=c \neq l,c\in Y $ \} with 
$|S_c|>N/2$.
\end{defn} 

 Thus, if the majority of S disagrees with $h$ on the class of $\mathbf{x}_s$  we have a fooling example for S.  $h_1...h_N$ may be humans. Since the given A is non-negative we want all of our classifiers to be able to perform the classifications of $\mathbf{x}$ and the test set $D_T$. The classifiers should also be able to perform acceptably for the given metric, i.e., the metric m  should be above a certain given threshold A for the test set $D_T$ and classifier $h_i$, $i=1...N$.

 Notice that the dimensionality of the feature space can be different between different classifiers as long as the classifier is capable of classifying for different dimensions. The more data with the fooling example property we have for a given dataset  $D_s$, the more difficulty S has in identifying decisions from $h$ for this dataset.
 
\begin{defn}
 Hidden message:  Given two classifiers $h$,$h_s\notin S$, we define a hidden message for S to be an example $\mathbf{x}_s \in X$   such that $h(\mathbf{x}_s)=h_s(\mathbf{x}_s)=l\in Y$, and there exist a subset of S such that $S_h=\{h_i \in S | h_i(\mathbf{x}_s)=c \neq l,c\in Y $ \} with $|S_h|>N/2  $.
\end{defn} 

The hidden message property implies that  $h$ and $h_s$ agree on the class decision for the example $\mathbf{x}_s$  while the majority of S disagrees. For a given dataset $D_s$,  this property  measures the difficulty that S has to identify data points where $h$ and $h_s$ agree on, or from the perspective of $h$ and $h_s$ how easy it is for S not to notice examples of agreement. $h$ can be thought as having the role of teacher for the context of intepretation-based indirect knowledge transfer.

\subsubsection{Quantification of Difficulty}

We use the CIFAR10 dataset \cite{krizhevsky2014cifar} to evaluate  different student classifiers, as it is easily recognizable by humans, but not as easy as MNIST. We  exclude the non-deep learning methods from the previous experiment due to their inferior performance.

  We use a CNN  as $h_1$ which achieves an accuracy of 86.72\% on the test set. $h_1$ is used  as "Teacher" ($h$ in the property definitions) for two sets of classifiers $S_1$ and $S_2$. $S_1$ comprises  two CNNs: CNN2 which has a larger architecture (more layers, more activations) and  CNN3 which has a larger architecture, and with different activations (see Appendix for details). For $S_2$ we use an enseble of five humans. We train the classifiers with the original training data, and give the humans a sample  to get experience with the images.

  The accuracies for the test set are in the first column of Table \ref{table:Res_S1} and \ref{table:Res_S2}. Additionally, a CNN with exactly the same architecture to $h_1$ (called CNN1) will be used as $h_s$ for both $S_1$ and $S_2$.
 With AM we produce 100 images (10 for each class) that have $h_1$ logits of at least 99\% for the target class (denoted as IMGS). We calculate the percentage of images that  a classifier agrees on the decision of the class with $h_1$ on.
Following the procedure from Section \ref{eval1_2}, we synthesize 100 more images (denoted as IMGS+L).

\begin{figure}
\centering
\includegraphics[height=3cm,width=7cm]{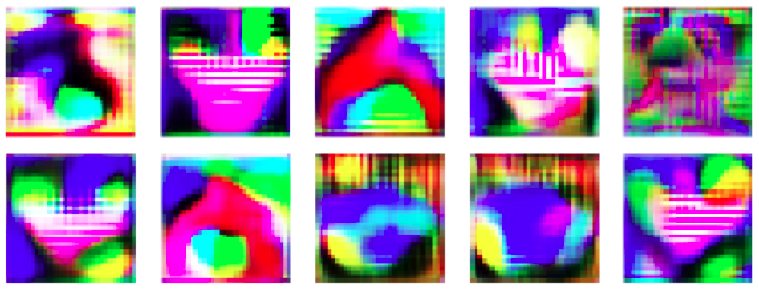}
\caption{\textit{Visualizations from the AM on CIFAR10 . }}
\label{fig:CIFAR10_1}
\end{figure}

 We calculate  the percentage of the dataset which satisfies the fooling example and hidden message properties. For  $S_1$ we use  the entire test set of CIFAR10 for this. Since humans would be overwhelmed to classify the entire CIFAR10 test set, we use 100 random subsamples from the test set for  $S_2$. The synthetic datasets also include only 100  images each, in order to be feasible for human classification.
 
\begin{table}

\caption{Results for $S_1$}
\label{table:Res_S1}
\vskip 0.15in

\begin{center}
\begin{small}
\begin{sc}

\begin{tabular}{lccccr}
\toprule
   &CIFAR10 Acc(\%)&imgs(\%) &imgs+L(\%) \\ 
\midrule
CNN1($h_s$)  &88.6 & 89.0 & 75.0 \\
\midrule
CNN2($\in S_1$)  & 90.65 & 75.0 & 51.0\\
 CNN3($\in S_1$)  &88.25 & 69.0 & 67.0\\
\midrule
 Hidden M.($S_1$)& 2.4 & 9.0& 11.0\\
 Fooling Ex.($S_1$)&  8.4 &11.0 & 21.0\\
\bottomrule
\end{tabular}%
\end{sc}
\end{small}
\end{center}
\end{table}

\begin{table}
\caption{Results for $S_2$. Using 100 images}
\label{table:Res_S2}
\vskip 0.15in
\begin{center}
\begin{small}
\begin{sc}

\begin{tabular}{lccccr}
\toprule
   $h\in$S2&sample Acc(\%)&imgs(\%) &imgs+L(\%) \\ 
\midrule
subject1  &86 & 26 & 12 \\
subject2  & 93 & 40 & 29 \\
subject3  &84 & 16 & 18\\
subject4  &89 & 20 & 27\\
subject5  &91 & 39 & 24\\
\midrule
 Fooling Ex.($S_2$)& 9 & 81& 89 \\
 Hidden M.($S_2$)& 5& 74&67 \\
\bottomrule
\end{tabular}%
\end{sc}
\end{small}
\end{center}
\end{table}

We see the  humans achieve a similar performance for the given sample as the CNNs (although the CNNs are evaluated with the whole dataset). Second, for IMGS and especially for IMGS$+$L,
the performance drops  for all  members of $S_1$ and $S_2$. Additionally, the percentage of examples that satisfy both properties increases. This means that the identification of what $h_1$ believes as classes for the examples of the datasets becomes increasingly  difficult to identify both for the individual members of the sets, and  for the combined decision of the sets. A very interesting point is that the human  ability to identify images in IMGS and IMGS$+$L is much lower than that of the members of $S_1$.

\subsection{Case Study: Creating Non-identifiable Training Data from  Apnea Recordings}
\label{eval_3}

Using ML techniques to automatically detect sleep apnea in arbitrary computer devices is hindered by the fact that sufficient trainining data is not available for many developers due to privacy and ownership issues. Therfore, we evaluate whether we can  benefit from the difficulty of interpreting the visualizations. On a subset of an open access apnea dataset called Apnea  ECG \cite{penzel2000apnea} we perform intepretation-based indirect knowledge transfer with $h_1$ and $h_2$ having  the same architectures. We train one network to distinguish between periods of normal breathing and periods with sleep apnea, and another network to distinguish between two possible teams of people from which the sleep recordings are obtained. We examine  the ability of student to learn from the visualizations, and the ability of the identifier network to distinguish between the two teams for the real and synthetic data.
We use full overnight sleep recordings from eight  patients. Every minute of the recording is labelled. We have two possible classes: an Obstructive Sleep Apnea (OSA) event happened or not.

\begin{table}
\vskip -0.2in
\caption{Apnea-ECG results  }
\vskip 0.1in
\label{table:Res_3}
\begin{center}

\begin{tabular}{ lcccr}
$h_2$ Acc(\%) & Base OSA & $h_2$ S. & $h_{idt}$ R.&$h_{idt}$ S. \\
\midrule

 Team 1 & 92.49  & 89.5 &91.07& 56.36\\
 Team 2 &86.69 & 86.10 &  89.98& 58.74\\
\bottomrule

\end{tabular}%
\end{center}
\vskip -0.25in
\end{table}

\par We separate these data into  training, validation, and  test sets via random subsampling and train a classifier $h_1$ to perform classification of OSA events per minute in the test set.
 Additionally, we train another classifier ($h_{idt}$) on the same training and test sets, but with the goal to identify  the person from which the data was recorded. This means that $h_{idt}$ outputs one of eight possible classes regarding the eight  patients. By using the procedure from Section \ref{method_3}, we generate a synthetic dataset, and we train a classifier $h_2$ with the same architecture as $h_1$. Finally, we evaluate whether $h_{idt}$ can identify the recordings for the synthetic dataset.

To perform the experiments we need to refine the generation process. For the real data, we know which person the data originated from. But for the generated data, we do not have access to the labels of the person which the generated example corresponds to.  Thus, we cannot measure the performance of $h_{idt}$ on the synthetic data.
To avoid this issue, we split the recordings into two teams of four people and follow the same procedure as before for each team separately. Then we train $h_{idt}$ to identify the team from which the data originates instead of the individual.

\par Table \ref{table:Res_3} presents our results. All the results regard the real test set. The second column (Base OSA) shows the results for a classifier trained on the real data on Team1 or Team 2 and tested on the real test set. The third column ($h_2$ S. ) denotes a classifier $h_2$ with identical architecture and activations to the original, trained with the examples of the learned features (the synthetic dataset) from the original for Team 1 or Team 2 which also performs classification on the test set. Columns 4 and 5 depict the accuracy of $h_{idt}$  in recognizing which team each example originates from for the real test data (Column 4) and for the synthetic data (Column 5).

There is a clear drop in the performance of $h_{idt}$ from the real to the synthetic data which means that $h_{idt}$ does not recognize as easily the team which the synthetic data was generated from as the team that data from the test set originate. Additionally  the synthetic data  can be used to train $h_2$ to achieve almost similar performance to $h_1$ (Base OSA) for both teams. 
 
\begin{figure}
\vskip -0.1in

\centering
\includegraphics[scale=0.15]{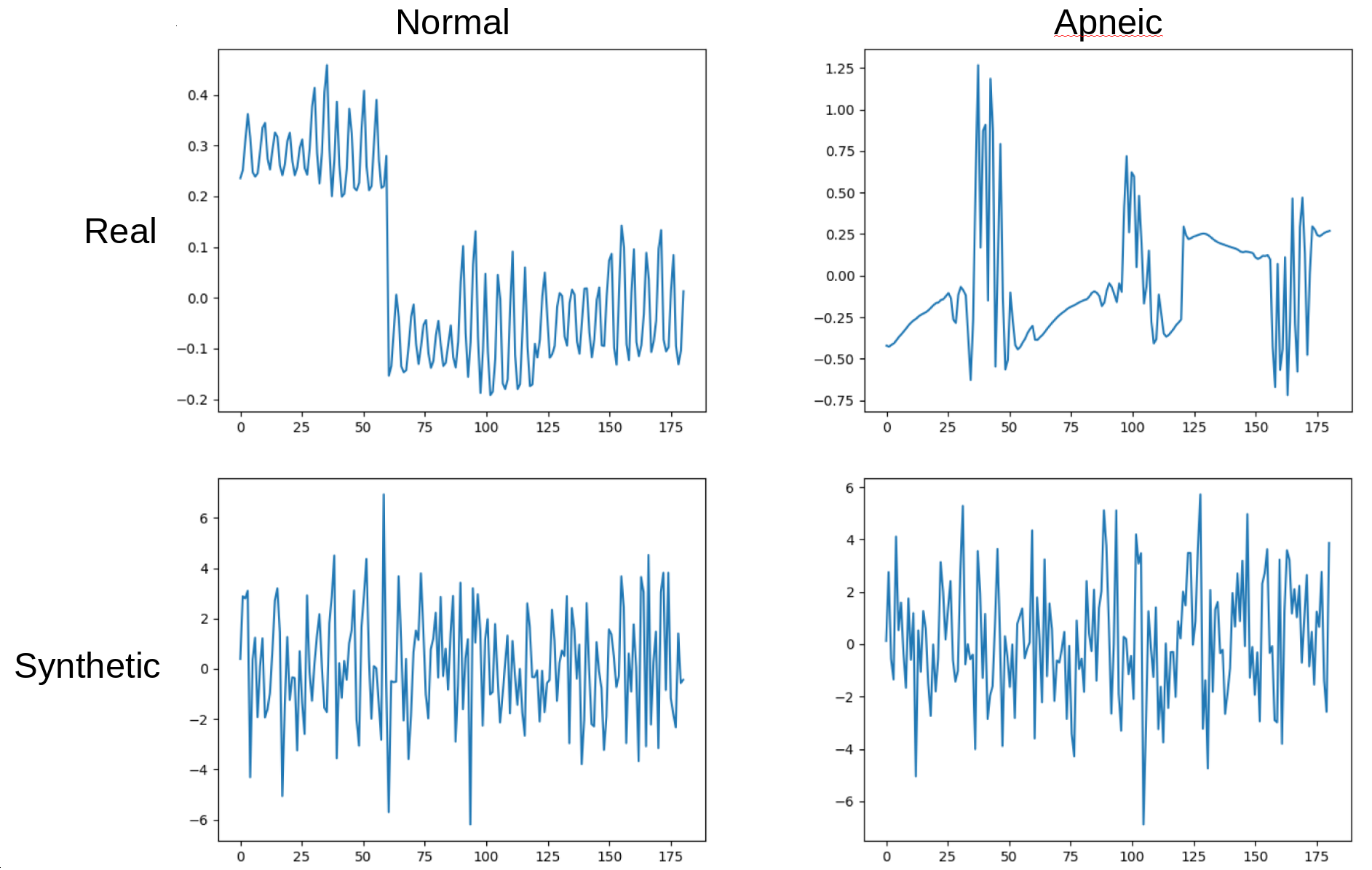}
\label{fig:Apnea_1}
\caption{ Example of real and generated Apnea-ECG data }
\vskip -0.1in
\end{figure}
\section{Conclusions}
\label{concl}

In this paper we propose an approach for interpretation-based indirect knowledge transfer between two classifiers $h_1$ and $h_2$. Using AM on the output logits of $h_1$,  G can interpret the knowledge that $h_1$ has about the distribution $p_{Data}$ of a dataset D, and generate visualizations of these interpretations. We then perform supervised learning of $h_2$ using only these visualizations. A primary benefit of this is that  $h_2$ can learn an approximation of $p_{Data}$ without access to the original dataset D. Furthermore, since the interpretation of G is guided only by the output logits of $h_1$, the resulting visualizations primarily contain features that are important to identify the target class. An added benefit is therefore that other, potentially sensitive information in $D$ is neglected from the visualizations. To strengthen this aspect, we extend G with a loss function $Loss_2$ that explicitly penalizes visualizations that contain high-level features of $D$.

We evaluate our approach experimentally with both computational and human classifiers. When $h_1$ and $h_2$ have similar architectures and algorithms (e.g., CNN of similar size), $h_2$ can successfully be trained using only the visualizations (i.e., without D). We achieve an accuracy of up to 95.51\% (for the same size of dataset as the real) on MNIST    and 89.5\% on Apnea-ECG. Interestingly, the accuracy drops gradually as the difference between $h_1$ and $h_2$ increases. This is particularly visible with human classifiers with a maximum accuracy of 40\% and 29\% with and without $Loss_2$, respectively. We develop new metrics to quantify this difficulty of learning, i.e., the fraction of points in the feature space that have one of two key properties: a fooling example is one that is mostly mis-classified by classifiers in a set S that  differ from $h_1$, and a hidden message is one that is  correctly classified by a classifier $h_s$ similar to  $h_1$, and is elsewise mostly mis-classified by S. With visualizations obtained using the CIFAR10 dataset, and where classifiers in S and $h_1$ are CNN that differ in size, 11\% and 18\% of the visualizations are fooling examples, with and without $Loss_2$, respectively, and 9\% and 14\% are hidden messages. When S contains humans  and $h_s$ is a CNN exactly the same in architecture to $h_1$, 81\% and 89\% are fooling examples and 74\% and 67\% are hidden messages. Our visualizations can be especially useful in a healthcare context where $D$ with sensitive information cannot be used directly. Using sleep recordings from the Apnea-ECG database, we demonstrate that: (1) our visualizations can be used to successfully train classifiers to detect sleep apnea, and (2) classifiers trained to identify groups of individuals in $D$ are incapable of discriminating between these groups in the visualizations.

As future work we plan to reduce  the time it takes to synthesize visualizations and the number of examples necessary to achieve  good performance. As a next step we aim to investigate whether we can leverage interpretation-based knowledge transfer to realize differetial privacy for sleep apnea data.

\bibliography{references}
\bibliographystyle{plain}

\clearpage

\appendix

\section{Soft Labels and Randomized T }

We evaluate the performance of CNN$_{ID}$ for different sizes of the dataset. Here instead of giving  a standard threshold T (as for example $T=0.99$), each time we perform activation maximization, we randomly choose a different value of T between 0.90 and 0.99. When we implement this change, we get the following results (Table \ref{table:Res_App1}):

\begin{table}[h]
\caption{Results  T=0.90-0.99, hard labels}
\label{table:Res_App1}
\begin{center}
\begin{small}
\begin{sc}

\begin{tabular}{lcr}
\toprule
 Size (\# of examples ) & Acc CNN$_{ID}$ (\%)\\ 
\midrule
60000 examples & 95.74$\pm$0.15\\
120000  examples&96.47$\pm$0.03\\
180000 examples&96.65$\pm$0.17\\
\bottomrule
\end{tabular}%

\end{sc}
\end{small}
\end{center}
\end{table}
 
 and: a score of 96.17\% for CNN$_S$ for 180000  examples (visualizations).  We additionally ran the other classifiers, but  we did not observe big differences in their performance.
 
 After that we additionally used soft labels for the CNN$_{ID}$ and we got  the following results:

\begin{table}[h]
\caption{Results  T=0.90-0.99, soft labels}
\label{table:Res_App2}
\begin{center}
\begin{small}
\begin{sc}

\begin{tabular}{lcr}
\toprule
 Size (\# of examples ) & Acc CNN$_{ID}$ (\%)\\ 
\midrule
400 examples &82.04$\pm$1.03\\
4000  examples&94.93$\pm$0.28\\
45000 examples&97.18$\pm$0.11\\
60000 examples&97.90$\pm$0.01\\
120000 examples&98.13$\pm$0.03\\
\bottomrule
\end{tabular}%

\end{sc}
\end{small}
\end{center}
\end{table}

and for CNN$_S$ we experimented  with 120000 samples and got an accuracy of: 97.84$\pm$0.06.

\section{CIFAR10 synthetic test sets}

\begin{figure}
\centering
\includegraphics[scale=0.17]{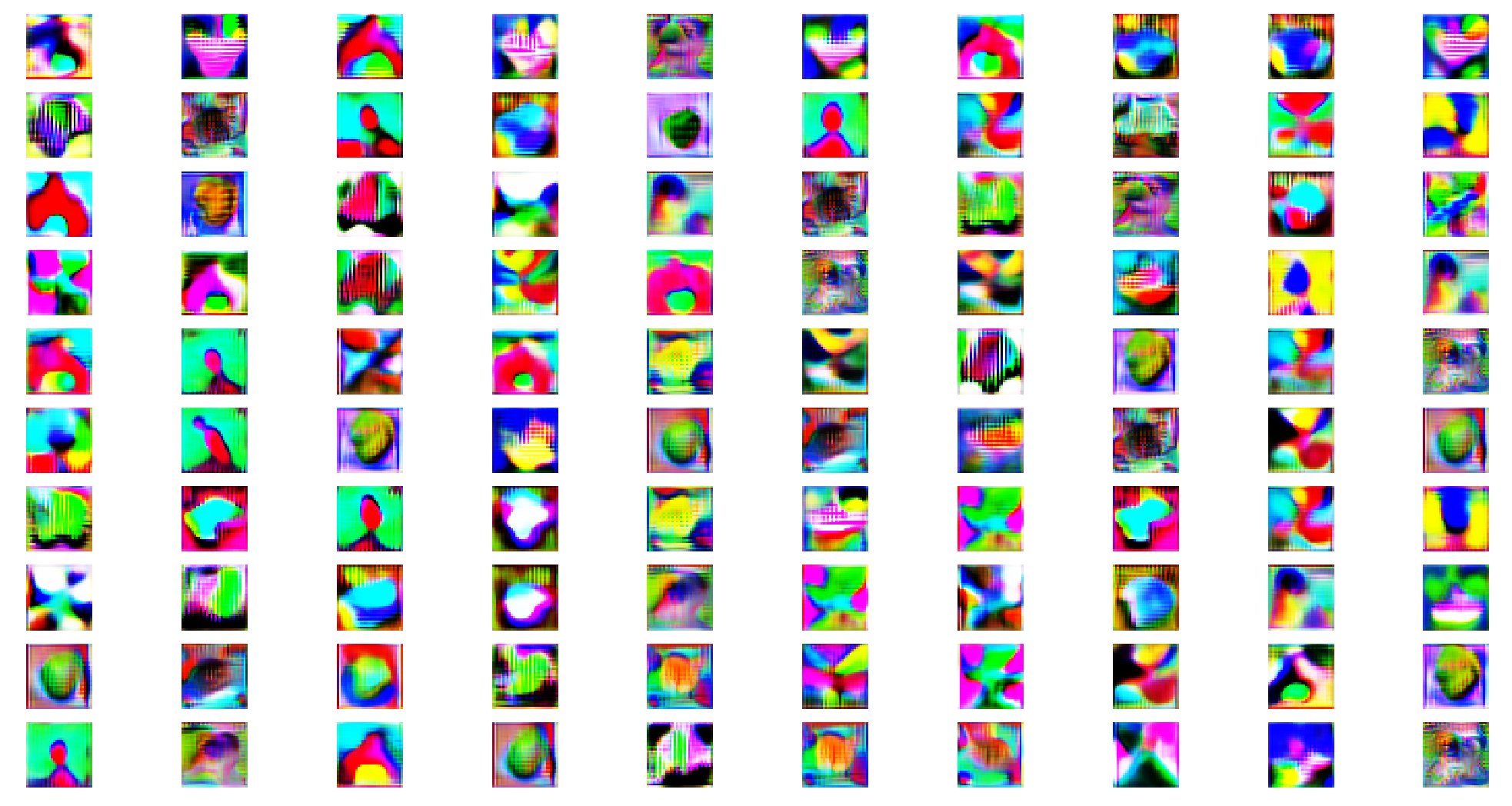}
\caption{\textit{ IMGS  }}
\label{fig:IMGS}
\end{figure}

\begin{figure*}
\centering
\includegraphics[scale=0.17]{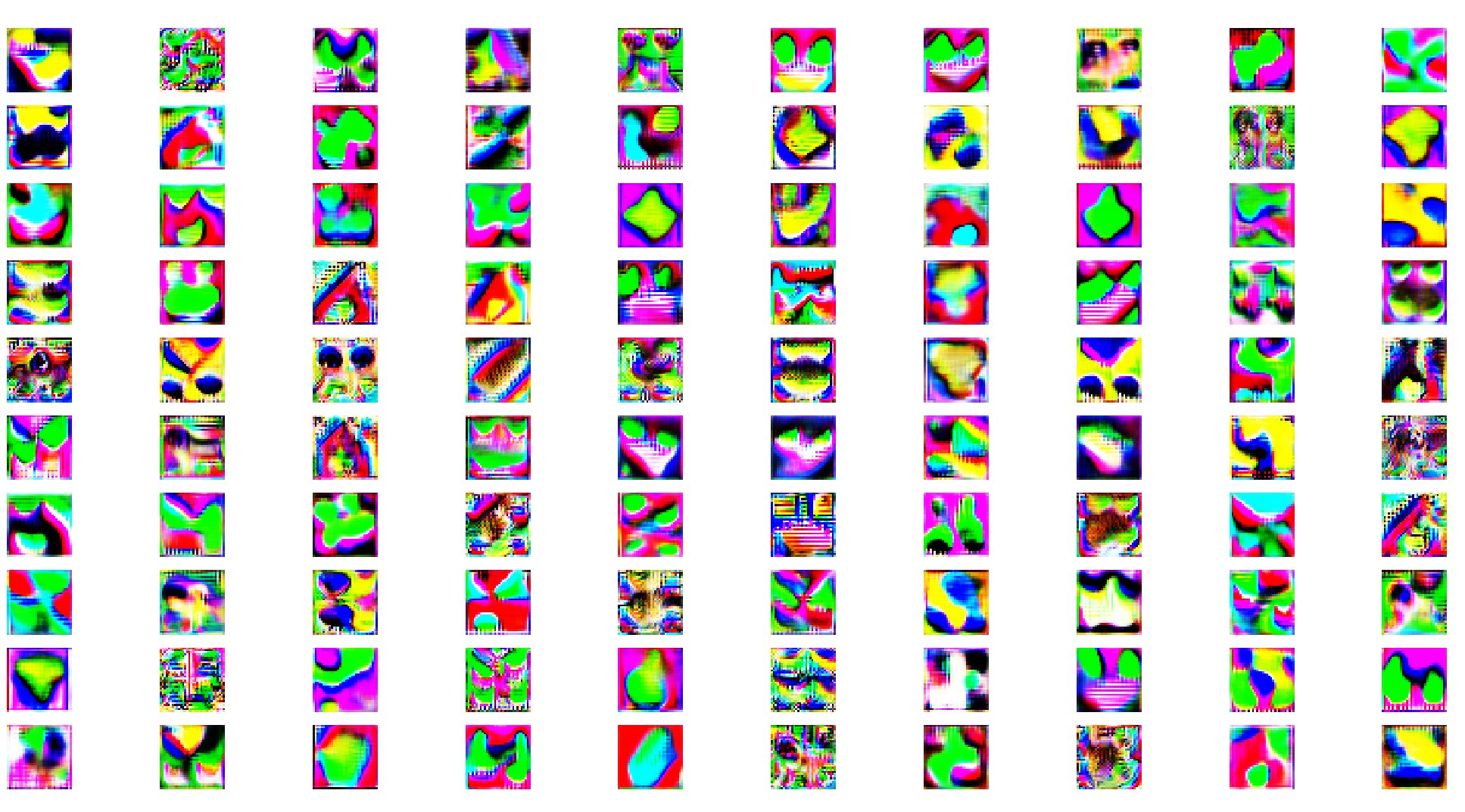}
\caption{\textit{ IMGS+L }}
\label{fig:IMGSL}
\end{figure*}

 IMGS and IMGS+L datasets (Figures \ref{fig:IMGS},  \ref{fig:IMGSL}).

\section{ Examples of MNIST visualizations}
\begin{figure}
\centering
\includegraphics[scale=0.35]{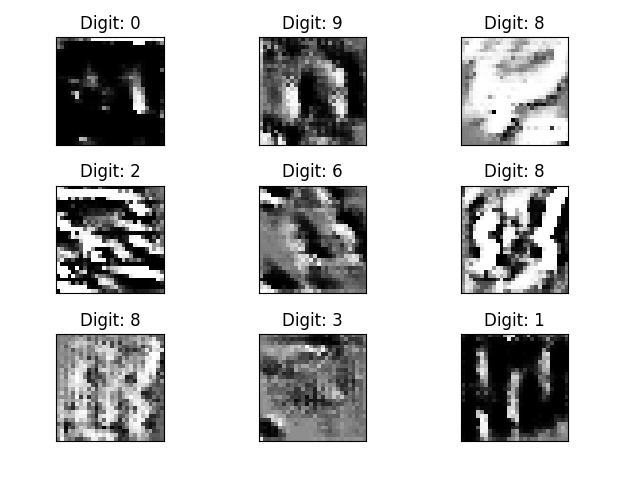}
\caption{\textit{Batch1  }}
\label{fig:IMGS1}
\end{figure}

\begin{figure}
\centering
\includegraphics[scale=0.35]{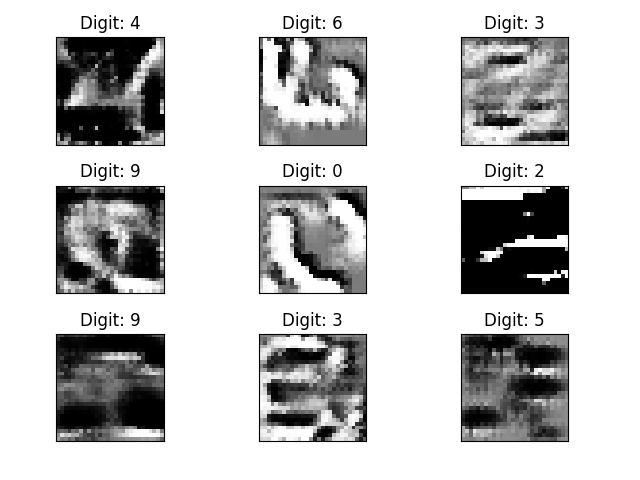}
\caption{\textit{ Batch2 }}
\label{fig:IMGS2}
\end{figure}

\begin{figure}
\centering
\includegraphics[scale=0.35]{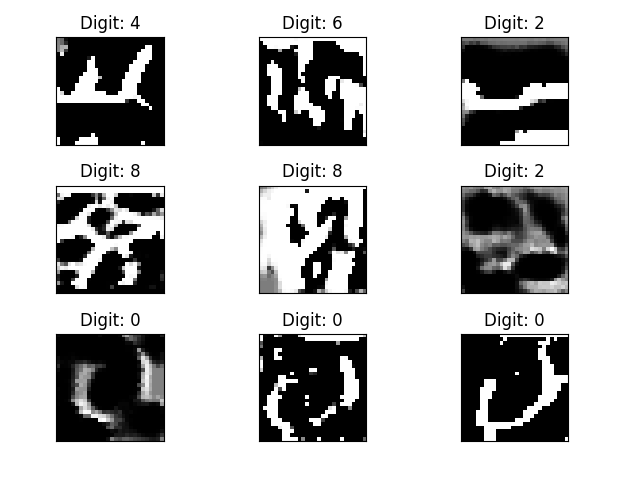}
\caption{\textit{Batch3  }}
\label{fig:IMGS3}
\end{figure}

Examples of visualizations of MNIST (Figures \ref{fig:IMGS1},\ref{fig:IMGS2},\ref{fig:IMGS3}). Each batch (9 imgs) corresponds to different pre-AM movement (Batch 1. Attraction toward one random class. Batch 2. Repulsion from  many classes Batch 3. Attraction to many classes). Notice that the movement is not the only difference between the Batches.  Parameters like learning rate and minibatch size also change per Batch.

\section{CIFAR10 Architectures of CNN1, CNN2 and CNN3}

For Experiment 2 (the CIFAR10 experiments) we wanted CNN2 and CNN3 that form $S_1$ to be at least  near equal in terms of performance on the original CIFAR10 test set to the performances of the CNN that acts as $h_1$ ( the "Teacher"-or h for the hidden message/fooling example property-called CNN$_{h_1}$) and of CNN1, which acts as  $h_s$ from the hidden message property. That way we have a stronger means of quantification   of the "difficulty" of assumingly progressively harder to interpret datasets (from CIFAR10 test set to IMGS and then to IMGS+L) via using $S_1$. 
General Architectures:
\begin{itemize}
\item CNN1=CNN$_{h_1}$: 6 convolutional 1 dense (elu activations) + batchnorm+dropout+data augmentation(ImagedataGenerator, rotation, horizontal flip width shift length shift)
\item CNN2: 10 conv 1 dense (elu activations)+ batchnorm+dropout+data augmentation(ImagedataGenerator, rotation, horizontal flip width shift length shift)
\item CNN3: 10 conv 3dense (relu activations)+ batchnorm+dropout+data (no augm)
\end{itemize}

\end{document}